\documentclass[journal]{IEEEtran}
\usepackage[cmex10]{amsmath}
\usepackage{times}
\usepackage[dvips]{graphicx} %%\usepackage{graphicx}
\usepackage{subfigure}
\usepackage{color}
\usepackage{multirow}

\usepackage{rotating}
\usepackage{bbm}
\usepackage{bm}
\usepackage{latexsym}

\DeclareMathOperator*{\argmax}{arg\!max}
\DeclareMathOperator*{\sign}{sign}

\ifCLASSINFOpdf
\else
\fi

\begin{document}
\title{Unbounded Output Networks for Classification}

\author{Stefan~Elfwing,
        Eiji~Uchibe,
        and~Kenji~Doya% <-this % stops a space
\thanks{This work was supported by project commissioned by the New Energy and Industrial Technology Development Organization (NEDO), MEXT KAKENHI grants  16H06563 and 17H06042, and Okinawa Institute of Science and Technology Graduate University research support to KD.}% <-this % stops a spa
\thanks{S. Elfwing, and E. Uchibe are with the Dept. of Brain Robot Interface, ATR Computational Neuroscience Laboratories, 2-2-2 Hikaridai, Seikacho, Soraku-gun, Kyoto 619-0288, Japan (E-mail: elfwing@atr.jp; uchibe@atr.jp).}% <-this % stops a spa
\thanks{K. Doya are with the Neural Computation Unit, 
Okinawa Institute of Science and Technology Graduate University,
 1919-1 Tancha, Onna-son, Okinawa 904-0495, Japan (e-mail: doya@oist.jp).}}% <-this % stops a space

\maketitle

\begin{abstract}
We proposed the expected energy-based restricted
Boltzmann machine (EE-RBM) as a discriminative RBM method for
classification. Two characteristics of the EE-RBM are that
the output is unbounded and that the target value of correct
classification is set to a value much greater than one. 
In this study, by adopting features of the EE-RBM approach to feed-forward neural networks, we propose the UnBounded output network (UBnet) which is characterized by three features: (1) unbounded output units; (2) the target value of correct classification is set to a value much greater than one; and (3) the models are trained by a modified mean-squared error objective.
We evaluate our approach using the MNIST, CIFAR-10, and CIFAR-100 benchmark datasets. We first demonstrate, for shallow UBnets on MNIST, that a setting of the target value equal to the number of hidden units significantly outperforms a setting of the target value equal to one, and it also outperforms standard neural networks by about 25\%. We then validate our approach by achieving high-level classification performance on the three datasets using unbounded output residual networks. We finally use MNIST to analyze the learned features and weights, and we demonstrate that UBnets are much more robust against adversarial examples than the standard approach of using a softmax output layer and training the networks by a cross-entropy objective.
\end{abstract}

\IEEEpeerreviewmaketitle

\section{Introduction}
\IEEEPARstart{W}{e} proposed the expected energy-based restricted Boltzmann machine (EE-RBM) as a discriminative RBM method for classification~\cite{Elfwing15}. The main difference between the EE-RBM architecture and the standard feed-forward neural network architecture is that the output is not computed in specific output nodes. Instead, the output is defined as the negative expected energy of the RBM, which is computed by the weighted sum of all bi-directional connections in the network. Two characteristics of the EE-RBM are that the output values are unbounded and that the target value of correct classification, $T$, is related to the size of the network and therefore set to a value much greater than one.  We have successfully applied the EE-RBM in the reinforcement learning domain~\cite{Elfwing16}, achieving what was then the state-of-the-art score in stochastic SZ-Tetris and achieving effective learning in a robot navigation task with raw and noisy RGB images as state input.

In this study, by adopting features of the EE-RBM approach to feed-forward neural networks, we propose the UnBounded output network (UBnet) which is characterized by three features:
\begin{enumerate}
\item Unbounded units in the output layer;
\item The target value of correct classification, $T$, is set to a value much greater than one;
\item The models are trained by a modified mean-squared error objective that gives more weight to errors that correspond to correct classification and less weight to errors that correspond to incorrect classification.
\end{enumerate}

We use the sigmoid-weighted Linear Unit (SiLU), which we originally proposed as an activation function for neural networks in the reinforcement learning domain~\cite{Elfwing18}. The SiLU unit is also based on the EE-RBM. The activation of the SiLU is computed by the sigmoid function multiplied by its input, which is equal to the contribution to the negative expected energy from one hidden unit in an EE-RBM, where the negative expected energy is equal to the negative free energy minus the entropy. 

We have successfully used the SiLU and its derivative (dSiLU) as activation functions in neural network-based function approximation in reinforcement learning~\cite{Elfwing18,Elfwing18b}, achieving the current state-of-the-art scores in SZ-Tetris and in $10\times10$-Tetris, and achieving competitive performance compared with the DQN algorithm~\cite{Mnih15} in the domain of classic Atari 2600 video games. After we first proposed the SiLU~\cite{Elfwing17a}, Ramachandran \emph{et al.}~\cite{Ramachandran17} performed a comprehensive comparison between the SiLU, the Rectified Linear Unit (ReLU)~\cite{Nair10}, and 6 other activation functions in the supervised learning domain. They found that the SiLU consistently outperformed the other activation functions when tested using 3 deep architectures on CIFAR-10/100~\cite{Krizhevsky09}, using 5 deep architectures on ImageNet~\cite{Deng09}, and on 4 test sets for English-to-German machine translation. 

We use the MNIST dataset to demonstrate that for shallow UBnets, a setting of $T$ equal to the number of hidden units significantly outperforms a setting of $T = 1$ and it also outperforms standard neural networks without additional optimization by about 25\%. We train \emph{UnBounded output Residual networks} (UBRnets) on the MNIST, CIFAR-10, and CIFAR-100 benchmark datasets and validate our approach by achieving high-level classification performance on the three datasets. We use the CIFAR-10 dataset to demonstrate a small but significant improvement in performance of UBRnets with SiLUs compared with UBRnets with ReLUs. We finally use MNIST to analyze the features and weights learned by UBnets and we demonstrate that UBnets are much more robust against adversarial examples~\cite{Szegedy13} than the standard approach of using a softmax output layer and training the networks by a cross-entropy objective.

\section{Method}
We proposed the EE-RBM~\cite{Elfwing15} as a discriminative
learning approach to provide a self-contained
RBM~\cite{Smolensky86,Freund92,Hinton02} method for classification. In an EE-RBM, the output $Q$ of an input vector $\mathbf{x}$ and
a class vector $\mathbf{y}_j$ (``one-of-$J$'' coding) is computed by
the negative expected energy of the RBM, which is given by the
weighted sum of all bi-directional connections in the network:
\begin{IEEEeqnarray}{rCl}
Q(\mathbf{x},\mathbf{y}_j) &=& \sum_k z_{kj}\sigma(z_{kj}) +
\sum_ib_ix_i + b_j, \\
 \sigma(x) &=& \frac{1}{1+e^{-x}}. 
\end{IEEEeqnarray}
Here, $z_{kj}$ is the input to hidden unit $k$ for class $j$, $b_i$ is
the bias weight for input unit $x_i$, and $b_j$ is the bias weight for
class unit $y_j$.

\begin{figure}[ht]
%% \vskip 0.2in
\begin{center}
\centerline{\includegraphics[width=0.9\linewidth]{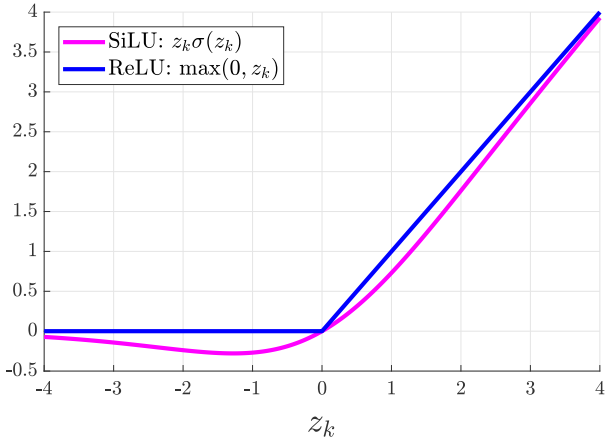}}
\caption{The activation functions of the SiLU ($z_k\sigma(z_k)$) and the
ReLU ($\max(0,z_k)$).}
\label{fig:EE_vs_relu}
\end{center}
%% \vskip -0.2in
\end{figure}

In this study, we use the SiLU for neural network-based classification. The 
activation of the SiLU is computed by the sigmoid function multiplied 
by its input, which is equal to the contribution to the output from 
one hidden unit in an EE-RBM. Given an input vector $\mathbf{x}$, 
the activation of a SiLU $k$ (in a hidden layer or the output layer), $a_k$, is given by:
\begin{IEEEeqnarray}{rCl}
a_k(\mathbf{x}) &=& z_k\sigma(z_k), \\
z_k(\mathbf{x}) &=& \sum_iw_{ik}x_i + b_k.
%% \sigma(x) &=& \frac{1}{1+e^{-x}}. 
\end{IEEEeqnarray}
Here, $w_{ik}$ is the weight connecting input $x_i$ and unit $k$, and
$b_k$ is the bias weight for unit $k$. For $z_k$-values of large
magnitude, the activation of the SiLU is approximately equal to the
activation of the ReLU (see Fig.~\ref{fig:EE_vs_relu}),
i.e., the activation is approximately equal to zero for large negative
$z_k$-values and approximately equal to $z_k$ for large positive
$z_k$-values. Unlike the ReLU (and other commonly used activation
units such as sigmoid and tanh units), the activation of the SiLU
is not monotonically increasing. Instead, it has a global minimum
value of approximately $-0.28$ for $z_{k} \approx -1.28$.

The derivative of the activation function of the SiLU, used for
stochastic gradient-descent updates of the weight parameters, is given by
\begin{equation}
\nabla_{w_{ik}} a_k(\mathbf{x}) = \sigma(z_k)x_i + \sigma(z_k)(1 -
\sigma(z_k))z_{k}x_i.
\end{equation}

Two features of the EE-RBM are that the output of the
network is unbounded and that the target value for correct
classification, $T$, is set to a value $\gg 1$. In this study, we
emulate this approach by proposing UBnets with unbounded units, such as the SiLU, in the output layer of standard feed-forward neural network architectures. The learning is achieved by a modified mean-squared error training objective:
\begin{IEEEeqnarray}{rCl}
J(\boldsymbol{\theta}) &=& 
\frac{1}{2N}\sum_{n}\sum_j\frac{1}{T}\left(Tt_j^n - y_j(\mathbf{x}^n)\right)^2 \label{Eq:J1}
 \\ 
&=& \frac{1}{2N}\sum_{n}\sum_jT\left(t_j^n -\frac{y_j(\mathbf{x}^n)}{T} \right)^2 \label{Eq:J2}.
\end{IEEEeqnarray}
Here, $t_j^n$ is the standard target value ($t_j^n = 1$ if training
example $\mathbf{x}^n$ belongs to class $j$, otherwise $t_j^n = 0$)
and $y_j$ is the output value for class $j$. The stochastic
gradient-descent update of the parameters, $\boldsymbol{\theta}$, for
an input vector $\mathbf{x}$ is then computed by either 
\begin{equation}
  \boldsymbol{\theta} \gets \boldsymbol{\theta} +
  \alpha \frac{1}{T}\sum_{j}\left[ Tt_j - y_j(\mathbf{x})  \right]
  \nabla_{\boldsymbol{\theta}}y_j(\mathbf{x})
  \label{eq:agg_upd}
\end{equation}
or
\begin{equation}
  \boldsymbol{\theta} \gets \boldsymbol{\theta} + \alpha
  \sum_{j}\left[t_j -\frac{y_j(\mathbf{x})}{T} \right]
  \nabla_{\boldsymbol{\theta}}y_j(\mathbf{x}).
\label{eq:udate2}
\end{equation}
Here, $\alpha$ is the learning rate. 

% We propose that using unbounded units in the output layer and a setting of $T \gg 1$ can produce robust and high-level performance classifiers.
% improves classification performance compared to both the setting of $T
% = 1$ and standard approaches, such as using softmax units in the
% output layer. 
For $T > 1$, the modified objective is not proportional
to the standard mean-squared error training objective. Errors
corresponding to incorrect classification ($t_j = 0$) are weighted
less, by a factor $1/T$, because $\left(T\cdot 0 - y_j\right)^2/T =
y_j^2/T$ (see (\ref{Eq:J1})). Errors corresponding to correct classification ($t_j = 1$) are weighted more. This is especially the case in the beginning of the
learning (assuming that the output weights are initialized so that the
outputs are zero or close to zero) when $y_j \ll T$ and $T\left(1 - y_j/T\right)^2 \approx T$ (see (\ref{Eq:J2})).

For negative input values to the output layer, $z_j < 0$, the output
$y_j$ is either equal to zero (ReLU output) or not a
monotonically increasing function (SiLU output). We,
therefore, classify input vectors with unknown class labels, $j^*$, in
the validation and test sets according to largest $z_j$-value:
\begin{equation}
j^* = \argmax_j^{}z_j.
\end{equation}

\section{Experiments}
We evaluated our approach using the MNIST~\cite{LeCun98}, CIFAR-10~\cite{Krizhevsky09}, and CIFAR-100~\cite{Krizhevsky09} benchmark datasets. We first used shallow UBnets (i.e., with one hidden layer) on the MNIST dataset to demonstrate that for networks with unbounded output units classification performance is significantly improved by setting the target value for correct classification $T \gg 1$. 

\begin{table}[htb!]
\renewcommand{\arraystretch}{1.3}
\caption{UBRnet architecture used for the MNIST, CIFAR-10, and CIFAR-100 experiments}
\label{tab:UBRnet}
\begin{center}
% \begin{sc}
\begin{tabular}{l|c}
Layer name & Layer type \\% \rule[-2ex]{0pt}{5ex} \\
\hline
\hline
inputBN & BN  \\%\rule[-2ex]{0pt}{5ex} \\
\hline
conv1 & conv: $3 \times 3 \times d$, BN \\%\rule[-2ex]{0pt}{5ex}\\
\hline
conv2\_x & $\left[ \begin{array}{c} 
\textrm{conv: } 3 \times 3 \times d\textrm{, BN} \\ 
\textrm{conv: } 3 \times 3 \times d\textrm{, BN} \\ 
\end{array} \right] \times n$ \rule[-4ex]{0pt}{9ex}\\
\hline
conv3\_x & $\left[ \begin{array}{c} 
\textrm{conv: } 3 \times 3 \times 2d\textrm{, BN} \\ 
\textrm{conv: } 3 \times 3 \times 2d\textrm{, BN} \\ 
\end{array} \right] \times n$ \rule[-4ex]{0pt}{9ex}\\
\hline
conv4\_x & $\left[ \begin{array}{c} 
\textrm{conv: } 3 \times 3 \times 4d\textrm{, BN} \\ 
\textrm{conv: } 3 \times 3 \times 4d\textrm{, BN} \\ 
\end{array} \right] \times n$ \rule[-4ex]{0pt}{9ex} \\
\hline
{} & max pooling $3 \times 3$, stride 2 \\%\rule[-2ex]{0pt}{5ex} \\
\hline
fc1 & fc: 2048 \\%\rule[-2ex]{0pt}{5ex} \\
\hline
fc2 & fc: \#classes \\%\rule[-2ex]{0pt}{5ex} \\
\hline
\end{tabular}
% \end{sc}
\end{center}
\end{table}
In the subsequent experiments, we used a residual network (ResNet)~\cite{He16} architecture with unbounded output (UBRnets; for details, see Table~\ref{tab:UBRnet}), similar to the architecture of the ResNets used in the CIFAR-10 experiments in~\cite{He16}. Our UBRnet architecture consisted of applying batch normalization (BN)~\cite{Ioffe15} to the input to the network, followed by a convolutional layer, three stacks of $n$ residual units, a max pooling layer, a fully-connected (fc) layer, and an unbounded output layer with either 10 (MNIST and CIFAR-10) or 100 (CIFAR-100) units. All convolutional filters were of size $3\times3$. 
%% Rev 7
There were $d$ convolutional filters in the first convolutional layer and the first stack of residual units (conv1 and conv2\_x in Table~\ref{tab:UBRnet}).
%% ENd rev 7
The number of convolutional filters were then increased by a factor of 2 in each stack of residual units after the first and, at same time, downsampling by a factor of 2 were performed using a stride of 2 in the first convolutional layer in the first residual unit (conv3\_1 and conv4\_1 in Table~\ref{tab:UBRnet}). All shortcut connections performed parameter free (option A in ~\cite{He16}) identity mappings~\cite{He16b}. Based on preliminary CIFAR-10 experiments, we changed the order of the modules in the residual units from convolution-BN-activation in the original residual unit~\cite{He16} to convolution-activation-BN (previously investigated in~\cite{Mishkin16}).

If not otherwise noted, the UBRnets were trained using a mini-batch size, $m$, of 100. The network weights were initialized as in ~\cite{He15}, except for the fully-connected layers, which were initialized using a zero-mean Gaussian distribution with standard deviation of 0.1 (first fc layer) and of 0.00001 (second fc layer). We did not use common optimization/regularization techniques such as momentum, weight decay, and dropout. Following \cite{Srivastava15} and \cite{He16}, we report test set performance as \emph{best (mean $\pm$ standard deviation)} based on five independent runs where the UBRnets were trained on the original training sets. The UBRnets were implemented using the MatConvNet toolbox for MATLAB~\cite{Vedaldi15}.

\subsection{MNIST}
\begin{figure}[ht]
%% \vskip 0.2in
\begin{center}
\centerline{\includegraphics[width=0.9\linewidth]{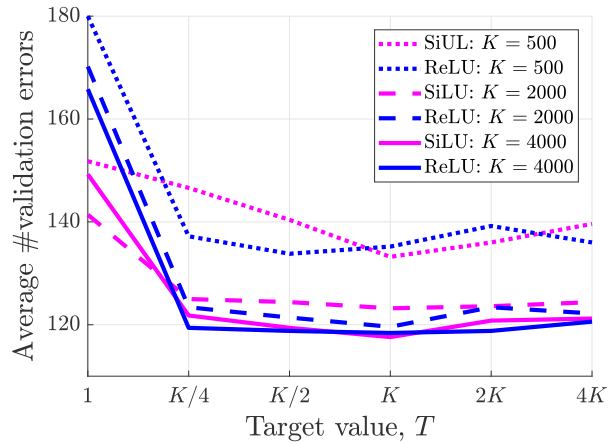}}
\caption{The average number of validation set errors (over 5
  experiments) on the MNIST dataset as a function of the target value
  for correct classification, $T$, for shallow SiLU and ReLU
  UBnets with 500, 2000, and 4000 hidden units, $K$.}
\label{fig:evalT}
\end{center}
%% \vskip -0.2in
\end{figure} 
The MNIST dataset~\cite{LeCun98} consists of 60000 training images and
10000 test images of ten handwritten digits, zero to nine, with an
image size of $28\times 28$ pixels. The grayscale pixel values were
normalized to the range $[0;1]$ by dividing the values by 255.

\begin{table}[htb!]
\renewcommand{\arraystretch}{1.3}
\caption{Test set error rate on the MNIST dataset without data augmentation.}
\label{tab:resMNIST}
\begin{center}
% \begin{sc}
\begin{tabular}{l|l}
Method & Test error rate (\%) \\
\hline
\hline
Maxout~\cite{Goodfellow13} & 0.45 \\
CKN~\cite{Mairal14} & 0.39 \\
DSN~\cite{Lee15} & 0.39  \\
FitNet-LSUV-SVM~\cite{Mishkin16} & 0.38\\
RCNN~\cite{Liang15} & 0.31 \\
UBRnet          & 0.28 (0.34 $\pm$ 0.039) \\
Gated Pooling~\cite{Lee16} & 0.29 $\pm$ 0.016 \\ 
MIN~\cite{Chang15} & \bf{0.24} \\
\end{tabular}
% \end{sc}
\end{center}
\end{table}
We first test our hypothesis that for UBnets classification performance is significantly improved by setting the target value for correct classification $T \gg 1$. We trained shallow UBnets with either SiLUs and ReLUs in both the hidden and the output layers, with 500, 2000, and 4000 hidden units ($K$) and 6 different settings of $T$: 1, $K/4$, $K/2$, $K$, $2K$, and $4K$. For each setting of $K$
and $T$, we trained the networks for 50 epochs and repeated each
experiment five times. The results show a quite remarkable improvement
for settings of $T \gg 1$ (see Fig.~\ref{fig:evalT}). For example,
in the experiments with 4000 hidden units, a setting of $T = 4000$
reduced the average number of validation set errors by 31 (from 149 to
118) for the SiLU UBnet and by 48 (from 166 to 118) for the ReLU
UBnet, compared with a setting of $T = 1$. 

\begin{table*}[htb!]
\renewcommand{\arraystretch}{1.3}
\caption{Test set error rate (\%) on CIFAR-10 and CIFAR-100 with standard data augmentation.}
\label{tab:resCIFAR10}
\begin{center}
% \begin{sc}
\begin{tabular}{l|l|l|l}
 Method & Depth & CIFAR-10 & CIFAR-100\\
\hline
\hline
DSN~\cite{Lee15} & - &  7.97 & 34.57  \\
All-CNN~\cite{Springenberg14} & - & 7.25 & 33.71 \\
Highway~\cite{Srivastava15} & - & 7.54 & 32.24 \\
ELU-CNN~\cite{Clevert16} & - & 6.55 & 24.28 \\
\hline
FractalNet~\cite{Larsson17} & 20 &  5.22 & 23.30 \\ 
with dropout/drop-path     & 20 &  4.60 & 23.73 \\ 
\hline
ResNet~\cite{He16} & 110 & 6.43 & 27.22~\cite{Huang16}\\ 
\hline
ResNet with & 110 & 5.25 & 24.98 \\
stoch. depth~\cite{Huang16} & 1202 & 4.91 & - \\
\hline
ResNet (pre-activation)~\cite{He16b} & 164  & 5.46 & 24.33 \\
{}                      & 1001 & 4.62 & 22.71 \\
\hline
Wide ResNet~\cite{Zagoruyko16} & 16 & 4.56 & 21.59 \\
                               & 28 & 4.17 & 20.43 \\
\hline
DenseNet-BC~\cite{Huang17}     & 100 & 4.51 & 22.27 \\
{}                             & 190 & \bf{3.46} & \bf{17.18} \\                  
\hline                      
\hline
UBRnet ($d = 25$) &  15   &  8.37 (8.67 $\pm$ 0.24) & - \\
UBRnet ($d = 50$) &  15 &  6.80 (7.04 $\pm$ 0.18) & - \\
UBRnet ($d = 50$) & 15 & 6.17 (6.27 $\pm$ 0.11) & - \\
UBRnet ($d = 150$) & 15 & 5.67 (5.85 $\pm$ 0.14) & 26.42 (26.70 $\pm$ 0.37)\\
UBRnet ($d = 300$) & 15 & 5.33 (5.54 $\pm$ 0.19) & 24.54 (25.00 $\pm$ 0.31)\\
UBRnet ($d = 450$) & 15 & 5.25 (5.35 $\pm$ 0.08) &  22.94 (23.26 $\pm$ 0.26) \\
\end{tabular}
% \end{sc}
\end{center}
\end{table*}

For both types of
networks and all settings of $K$, a setting of $T = K$ achieved
slightly better, or equally good, average performance. The only
exception was the ReLU UBnet with 500 hidden units where $T = K/2$
performed slightly better. Based on these results, $T$ was set to the
number of hidden units in the last hidden layer in the subsequent experiments.

The result of about 120 errors is a large improvement compared with the approximately 160 errors achieved by standard neural networks with either sigmoid~\cite{Simard03} or ReLU~\cite{Srivastava14} hidden units in the permutation invariant version of the MNIST task that do not use dropout training or other advanced regularization/optimization techniques. 

On the MNIST dataset, we trained UBRnets with SiLUs for 25 epochs without data augmentation, using a fixed learning rate $\alpha = 0.01$, $n = 1$, and $d = 64$ (one residual unit in each of the 3 stacks with 62, 128, and 256 filters, see Table~\ref{tab:UBRnet}). The target value for correct classification $T$ was set to 2048, as there were 2048 SiLUs in the first fully-connected layer (see Table~\ref{tab:UBRnet}). As shown in Table~\ref{tab:resMNIST}, our UBRnet achieved a test error rate of $0.28\%\ (0.34\pm0.039\%)$, which is slightly worse than the current state-of-the-art of $0.24\%$ achieved by batch normalized maxout network in networks (MIN)~\cite{Chang15}. 

\subsection{CIFAR-10 and CIFAR-100}
The CIFAR-10 dataset~\cite{Krizhevsky09} consists of natural
$32\times32$ RGB images belonging to 10 classes. The training set
contains 50000 images and the test set contains 10000 images. We preprocessed the images by subtracting the per-pixel mean value, computed over the training set. We followed the standard data augmentation: four zero-valued pixels were padded to each side and a $32\times32$ crop was randomly sampled from the padded image or its horizontal flip. During testing, the original $32\times32$ images were evaluated. The networks were trained for 100 epochs and the learning rate $\alpha$ was annealed by a factor of $\sqrt{10}$ after every 40 epochs. The target value for correct classification $T$ was set to 2048, as there were 2048 SiLUs in the first fully-connected layer (see Table~\ref{tab:UBRnet}).

We first compared UBRnets with SiLU and ReLU activation functions in both the hidden and the output layers. We trained the UBRnets with $n=2$ and $d=150$, and performed 10 independent runs for each of two activation functions. The SiLU UBRnet achieved a mean test error rate of 5.85 $\pm$ 0.14\%, which is significantly better (Wilcoxon signed-rank test with $p < 0.01$) than the 6.04 $\pm$ 0.15\% achieved by the ReLU UBRnet.

Following the result in~\cite{Zagoruyko16} that showed that width of residual networks (i.e., the number of convolutional filters) is more important for performance than the depth (i.e., the number of layers), we trained SiLU UBRnets with $n = 2$ (15 layers) and $d = \{25,50,100,300,450\}$. In the case of $d = 450$, the mini-batch size, $m$, was halved to $50$. The results of the experiments, as well as the reported results of other methods, are summarized in Table~\ref{tab:resCIFAR10}~\footnote{The UBRnet result for $n=2$ and $d =150$ is based on 10, instead of 5, independent runs, as it is taken from the previous experiment comparing SiLUs and ReLUs}. The results shows a large effect of increasing the width of the UBRnets, decreasing the test error rate from 8.37\% for $d = 25$  to 5.25\% for $d = 450$.

\begin{figure*}[!th]
\begin{center}
\subfigure {
 \includegraphics[width=0.45\linewidth]{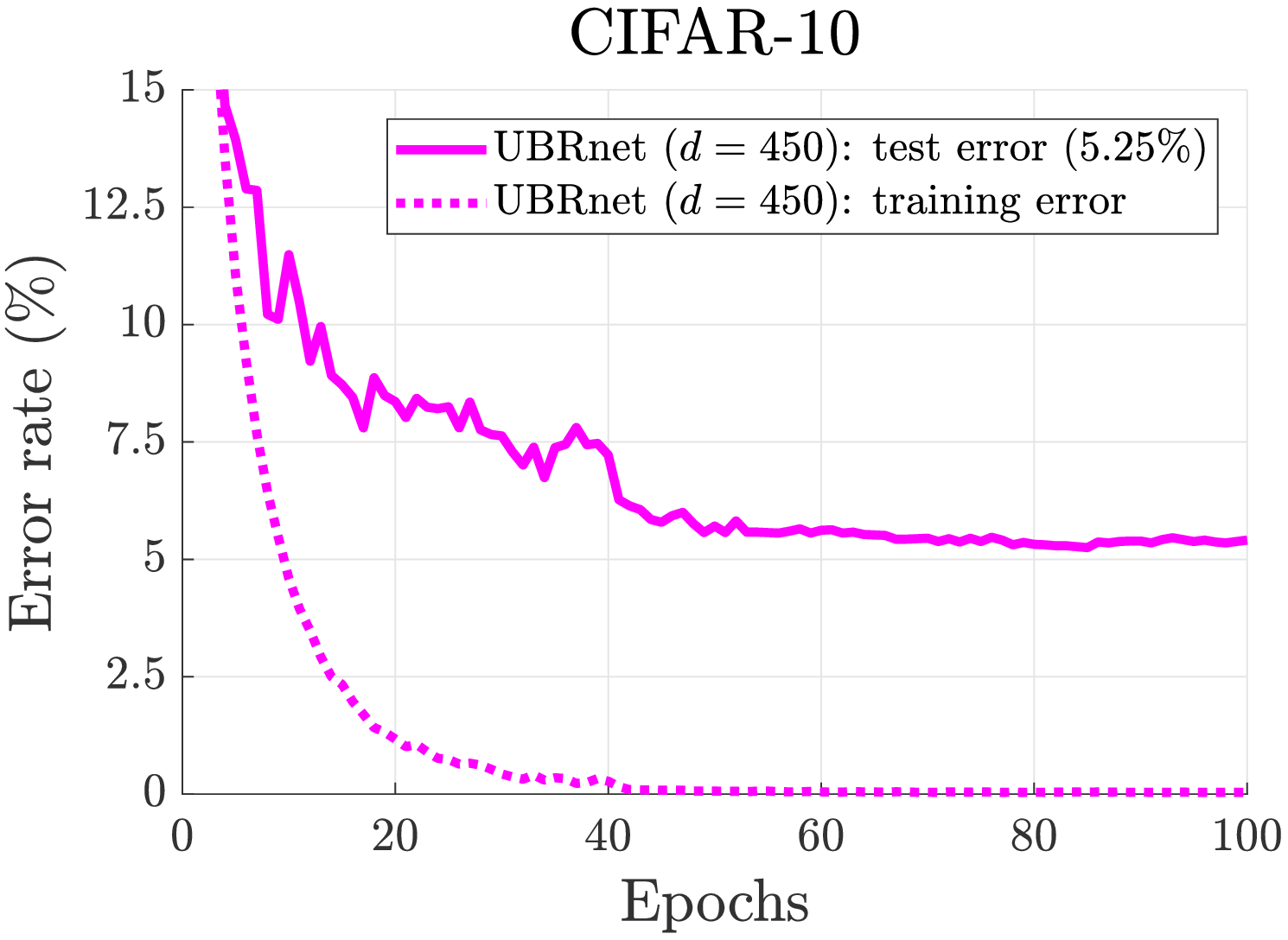}
}
\subfigure {
  \includegraphics[width=0.45\linewidth]{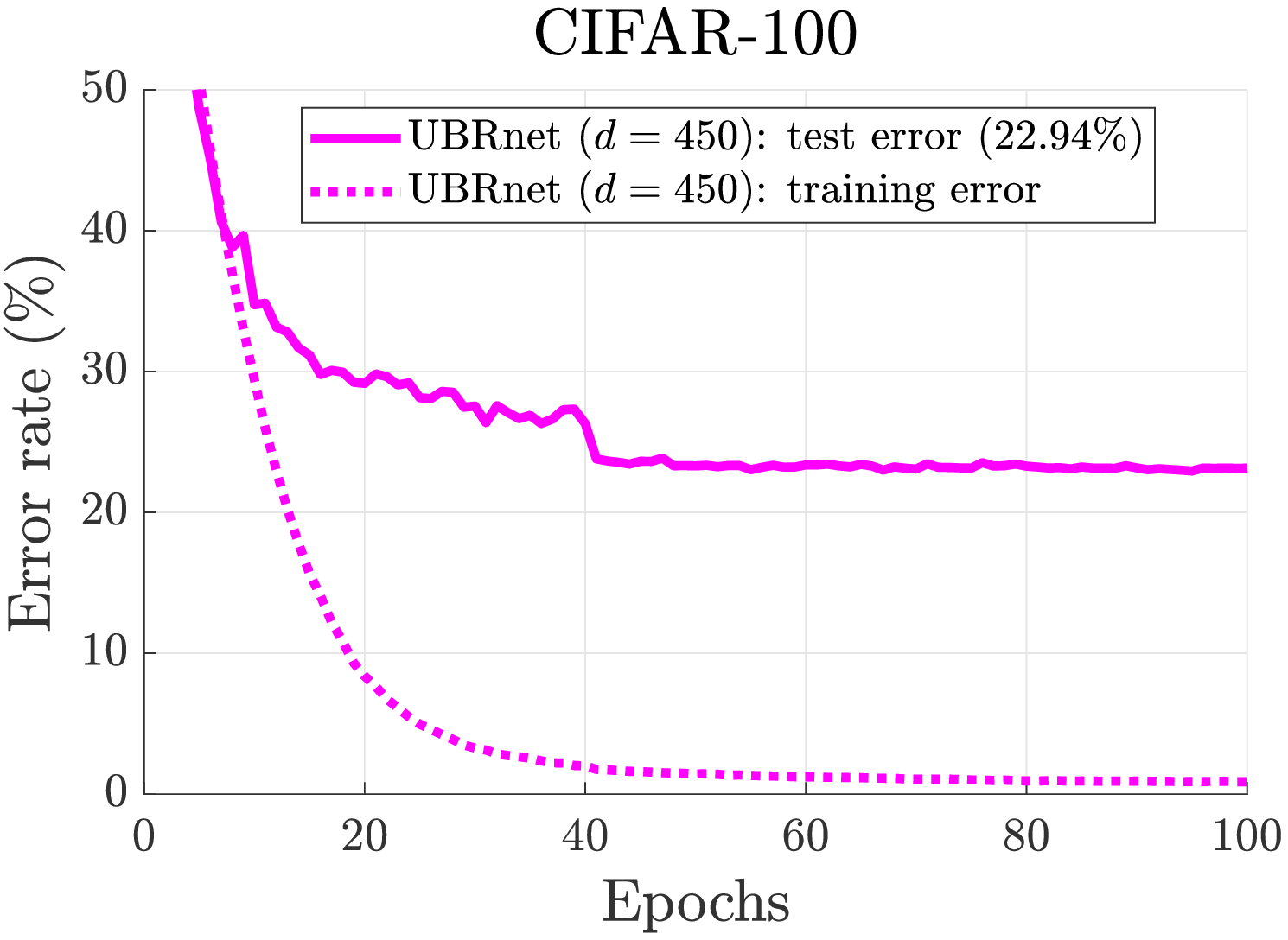}
} 
\end{center}
\caption{Learning curves for the best runs in CIFAR-10 (left) and CIFAR-100 (right) for the UBRnet with $d=450$. The solid lines show the test error rates and the dotted lines show the training error rate.}
\label{fig:resCIFAR}
\end{figure*}

The CIFAR-100 dataset~\cite{Krizhevsky09} has the same format and size as the CIFAR-10 dataset. The number of classes is 100, i.e., the number of training images per class is a tenth of the number in CIFAR-10. We used the same experimental setup (preprocessing, UBRnet architecture, meta-parameter settings, and data augmentation) as in the CIFAR-10 experiments. We trained UBRnets with $n=2$ and $d = \{150,300,450\}$. The test error rate (see Table~\ref{tab:resCIFAR10}) decreased from 26.42\% for $d=150$ to 22.94\% for $d=450$. In contrast with CIFAR-10, there was a large improvement in performance for increasing $d$ from $300$ (24.54\%) to 450 (22.94\%), which suggest that there is room for further improvement by further increasing the width of the UBRnet.

The classification results achieved by our UBRnet are encouraging, as they were achieved with minimal use of optimization/regularization techniques, and without using pre-activation in the residual units (i.e., batch normalization and the activation function are placed before each convolutional layer~\cite{He16b}). The UBRnet with $d=450$ outperformed the 164-layer ResNet with pre-activation and performed only slightly worse than the extremely deep 1001-layer ResNet. Our UBRnet results have only been surpassed by a large margin by the wide ResNet~\cite{Zagoruyko16} and the DenseNet~\cite{Huang17}, both of which used pre-activation.      

The learning was stable and fast as shown in Fig.~\ref{fig:resCIFAR}. For example, on CIFAR-10, the UBRnet with $d=450$ reached a 10\% test error rate after only 12 epochs. In comparison, the ResNets with pre-activation reached a 10\% test error rate after 40-50 epochs (estimated from Fig. 1 and 6 in~\cite{He16b}) and the DenseNet reached it after more than 50 epochs (right panel in Fig. 4 in~\cite{Huang17})

\section{Analysis of Unbounded Output Units}
A distinct feature of our approach of using unbounded units
in the output layers is that the value of an output unit $y_i$ (and
thereby the input to the output unit, $z_j$) for correct
classification is trained to match an exact value. The training error
($t_j - y_j/T$) can therefore be both positive and negative. In
contrast, for softmax output units, the training error is determined
by the differences in $z_j$-values and the training error is always
non-negative for correct classification. To investigate the effect of
our approach, we used the MNIST task and trained 3 shallow
network with 2000 hidden units for 50 epochs on the original training
set: a SiLU UBnet with $T=2000$ (SiLU-T2k), a SiLU UBnet with $T=1$ (SiLU-T1), and a network with SiLU hidden units and a softmax output layer that were trained with a cross-entropy
objective (SiLU-SM). They achieved the following test error rates (training error rate): 1.27\% (0.26\%) by the SiLU-T2k network, 1.49\% (0.14\%) by the SiLU-T1 network, and 1.64\% (0\%) by the SiLU-SM network.

\begin{figure*}[htb!]
\begin{center}
\includegraphics[width=1.0\linewidth]{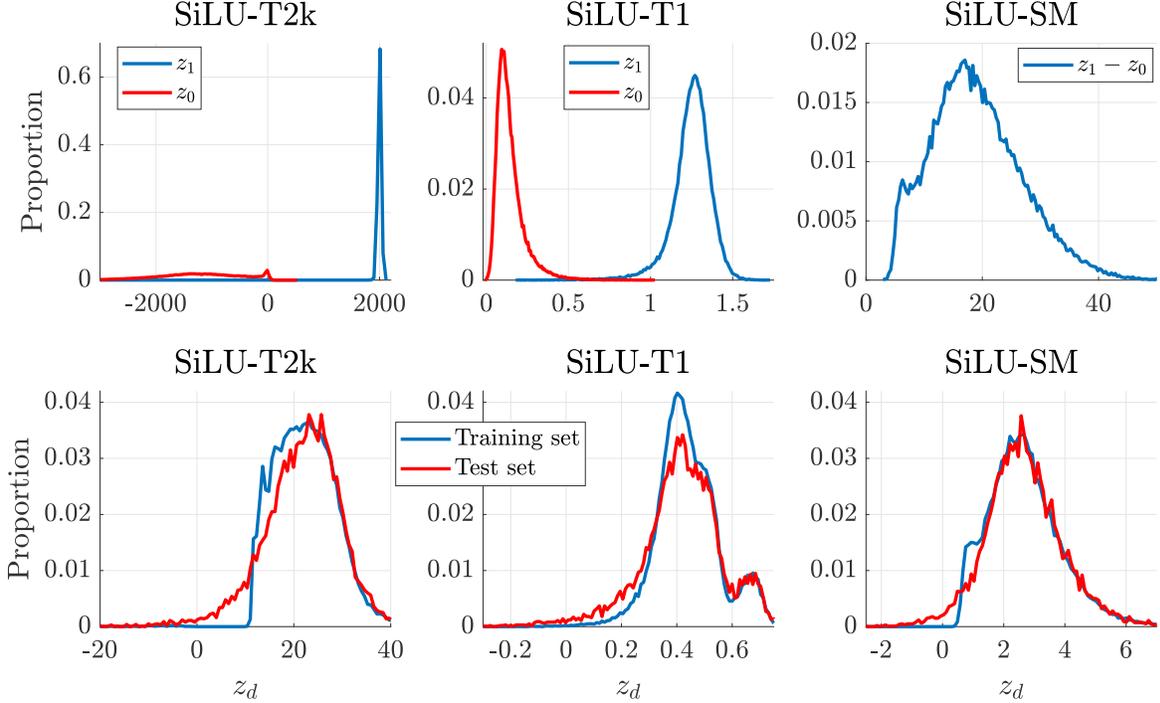}
\end{center}
\caption{Normalized histograms (150 bins) of the $z_1$- and
  $z_0$-values for the SiLU-T2k and SiLU-T1 networks and the $(z_1-z_0)$-values for the SiLU-SM network for the training set (top row), and normalized histograms of the  $z_d$-values (bottom row) for the training and test sets.}
\label{fig:dist_z0_z1_zd}
\end{figure*} 
To investigate the differences between the networks, we looked at the $z_j$-values after learning. If the correct class for an input vector is denoted $j^*$, let $z_1 =
z_{j*}$ (the $z_j$-value for correct classification) and $z_0 =
\max_{j\ne j^*} z_j$ (the maximum $z_j$-value for incorrect
classification). To get a measure of the networks' ability to separate
the $z_j$-values for correct and incorrect classification, we computed
the normalized distance between $z_1$ and $z_0$ (similar to margin
analysis for support vector machines): $z_d =
(z_1$-$z_0)/\|\mathbf{w}_{j^*}\|$, where $\mathbf{w}_{j^*}$ is the
weight vector incident on output unit $j^*$. Negative $z_d$-values
correspond to incorrect classified instances.

The distributions of the $z_1$- and $z_0$-values for the training set
(see the top row in Fig.~\ref{fig:dist_z0_z1_zd}) show distinct
differences between the networks. For the SiLU-T2k network, there was a clear separation between the narrow $z_1$-distribution with a peak at almost exactly $2000$ (mean value of $1999.4$) and the very wide
$z_0$-distribution with a mean value of about -1200. There was almost
no overlap between the two distributions, except for a small number of
images (157 or 0.26\,\%) that were not only wrongly classified, but
their $z_1$-values were negative and often of large magnitude (mean of
about -1400). For the SiLU-T1 network, there was considerable overlap
between the $z_1$-distribution with a peak at about 1.25 ($y_j \approx
0.97$) and the $z_0$-distribution with a peak at about 0.14 ($y_j
\approx 0.08$). This strongly suggests that the worse performance of
the SiLU-T1 network can be explained by that a target value of $1$ is
too small to learn a large enough separation of the $z_j$-values for
incorrect and correct classification. For the SiLU-SM network, there was
no wrongly classified training images, i.e., $z_1$-$z_0 > 0$ for all
images. However, for a relatively large number of images the
($z_1$-$z_0$)-values were relatively small, e.g., $z_1$-$z_0 < 10$ for
about 12\,\% of the training set. 

The bottom row in Fig.~\ref{fig:dist_z0_z1_zd} shows the
$z_d$-distributions for the training and test datasets. For the training set, the
$z_d$-values for the SiLU-T2k network were much larger than the other
two networks. For example, the minimum $z_d$-value for correct
classification was ~10.9, compared to ~0.45 for the SiLU-SM
network. 

% The experiments clearly shows that The larger ``safety
% margin'' of the SiLU-T2k network made it more robust to changes in the
% input images. Compared with the SiLU-SM network, the SiLU-T2k network
% reduced the number of errors between ~$23\,\%$ (test set) and ~$41\,\%$
% (thin images) on the four unseen datasets (see Table~\ref{tab:MNIST_trans}).

\subsection{Adversarial Examples}
Recent works~\cite{Biggio13,Szegedy13,Goodfellow15} have shown that 
neural networks are vulnerable to adversarial examples, i.e., they misclassify examples that are only slightly different than correctly classified examples. The changes can be so small that they are not visible to the human eye~\cite{Szegedy13}.
The larger ``safety margin'' of the SiLU-T2k network, in the form of larger $z_d$-values, suggests that it would be more resilient against adversarial examples. To test this hypothesis, we created adversarial examples of the MNIST test set using the fast gradient sign method~\cite{Goodfellow15}, where an adversarial example $\mathbf{x}_{\epsilon}$ is created from the original example $\mathbf{x}$ according to 
\begin{equation}
\mathbf{x}_{\epsilon} = \mathbf{x} + \epsilon\sign\left(\nabla_xJ(\boldsymbol{\theta},\mathbf{x})\right). 
  \label{eq:fast_sign_grad}
\end{equation}
\begin{figure}[!ht]
\begin{center}
\includegraphics[width=1.0\linewidth]{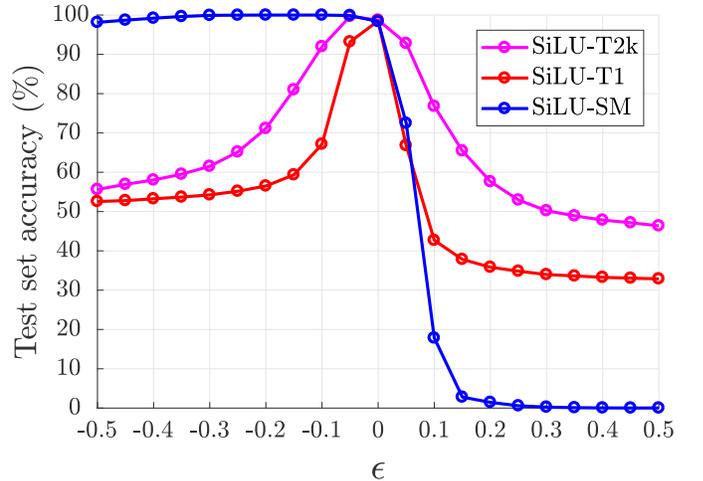}
\end{center}
\caption{Average test set accuracy of the SiLU-T2k, SiLU-T1, and SiLU-SM networks on the MNIST test set for adversarial examples with $\epsilon$ varied from $-0.5$ to $0.5$ with $0.05$ increments.}
\label{fig:adv_ex}
\end{figure}
The result of the experiment (Fig.~\ref{fig:adv_ex}) shows that the UBnets, and especially the SiLU-T2k network, were much more resilient against adversarial examples. For example, for $\epsilon = 0.1$, the networks achieved test set accuracy rates of 76.8\% (SiLU-T2K), 42.7\% (SiLU-T1), and 17.8\% (SiLU-SM). For $\epsilon = 0.25$, the SiLU-SM network could only correctly classify 58 test set images (0.58\%), which is similar to the accuracy rate of 0.1\% achieved by a shallow softmax network in ~\cite{Goodfellow15}. In the same study, a maxout network achieved an accuracy rate of 10.6\%, which  is much worse than the 52.94\% accuracy rate achieved by the SiLU-T2k network in this study. The SiLU-T2k network maintained an almost 50\% accuracy rate as $\epsilon$ increased to 0.5. 

% We created adversarial examples of the MNIST test set, varying $\epsilon$ from $-0.5$ to $0.5$ with $0.05$ increments. The results of the experiment 
% Fig.~\ref{fig:adv_ex} show the average test set accuracy of the three types of networks for the 21 different values of $\epsilon$.  

\subsection{Learned Features and Weights}
\begin{figure}[!ht]
\begin{center}
\subfigure {
 \includegraphics[width=1\linewidth]{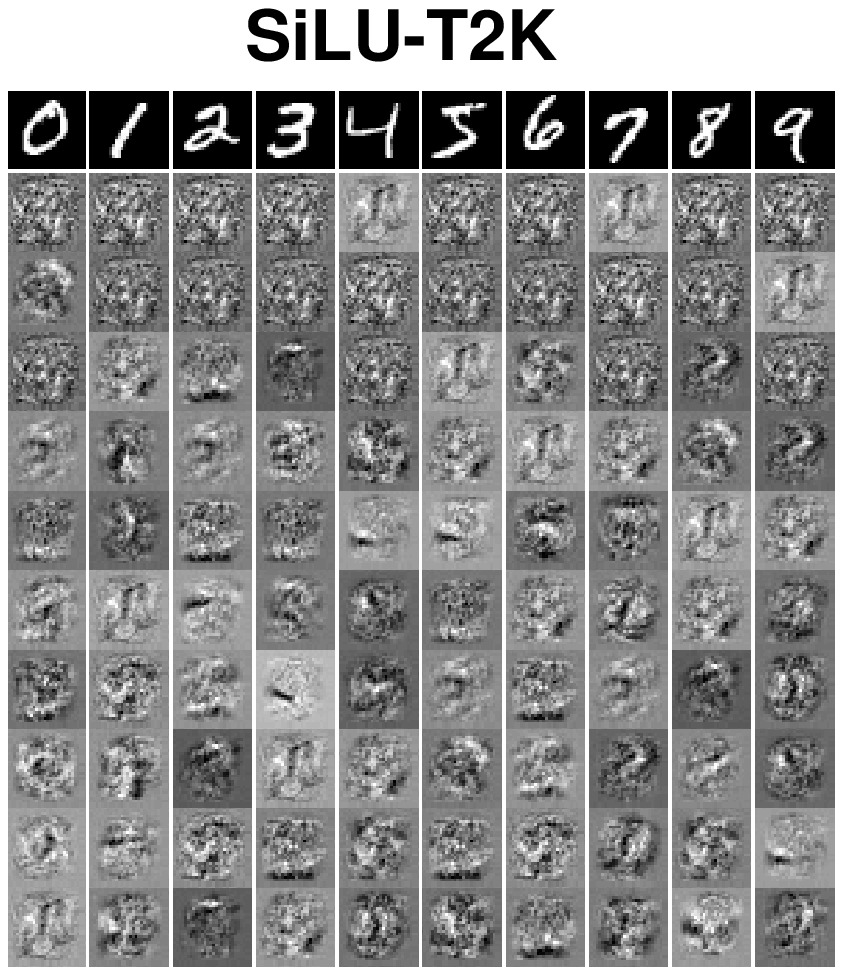}
}
\subfigure {
 \includegraphics[width=1\linewidth]{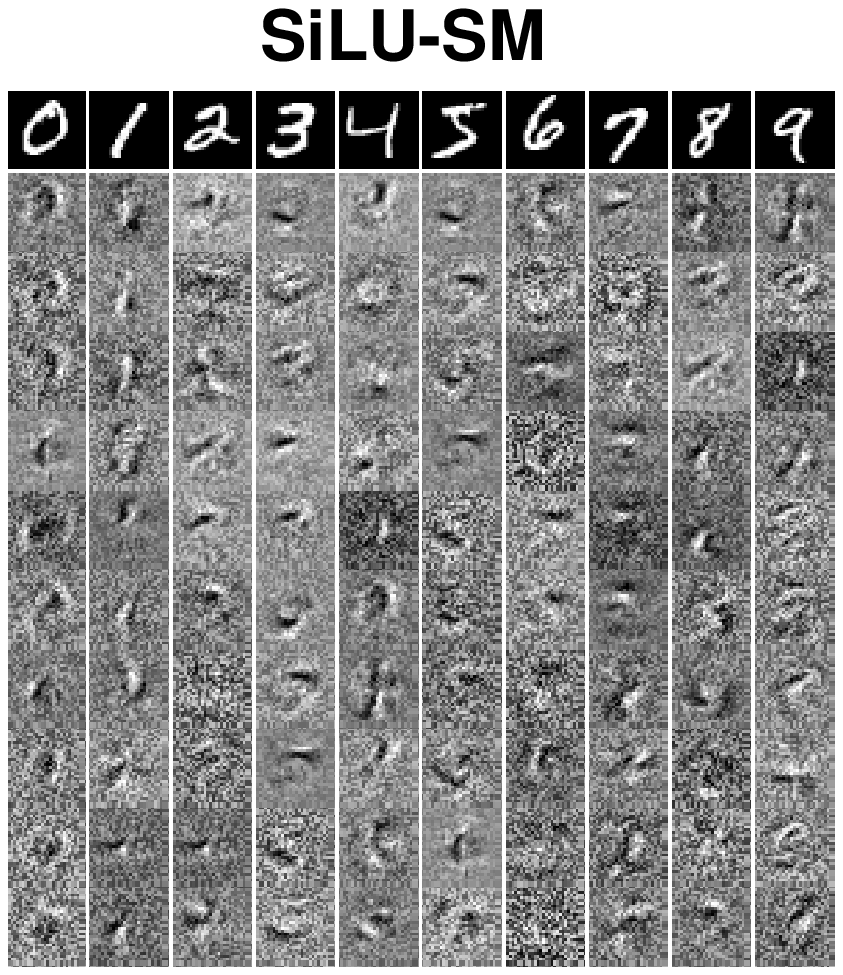}
} 
\end{center}
\caption{The learned hidden layer filters for the SiLU-T2k network (top
  panel) and the SiLU-SM network (bottom panel). The columns show the 10
  learned hidden layer filters with the highest median activation for
  each class (shown in order with the highest value at the top).}
\label{fig:filters}

\end{figure}

To investigate the difference in learned hidden layer feature
representation between the SiLU-T2k network and the SiLU-SM network, we
computed the median activation of each unit in the hidden layer for
100 randomly selected images from the training set from each class.
The columns of Fig.~\ref{fig:filters} show the 10 learned hidden
layer filters with the highest median activation for each class (shown
in order with the highest value at the top). The visualization shows a
clear difference between the two methods. In the SiLU-SM network, the
10 learned filters with the highest median activations were, to large
degree, different for each class. Only 14 filters were shared by two
classes and no filter was shared by more than two classes. In
contrast, in the SiLU-T2K network, there were 4 filters that were
shared by more than 7 classes. Two of the filters were shared by all
classes: the first filter generated the highest or second highest
median activation for all classes and the second filter generated the
second or third highest median activation for all classes.

\begin{figure}[!htb]
\begin{center}
\subfigure {
  \includegraphics[width=1.0\linewidth]{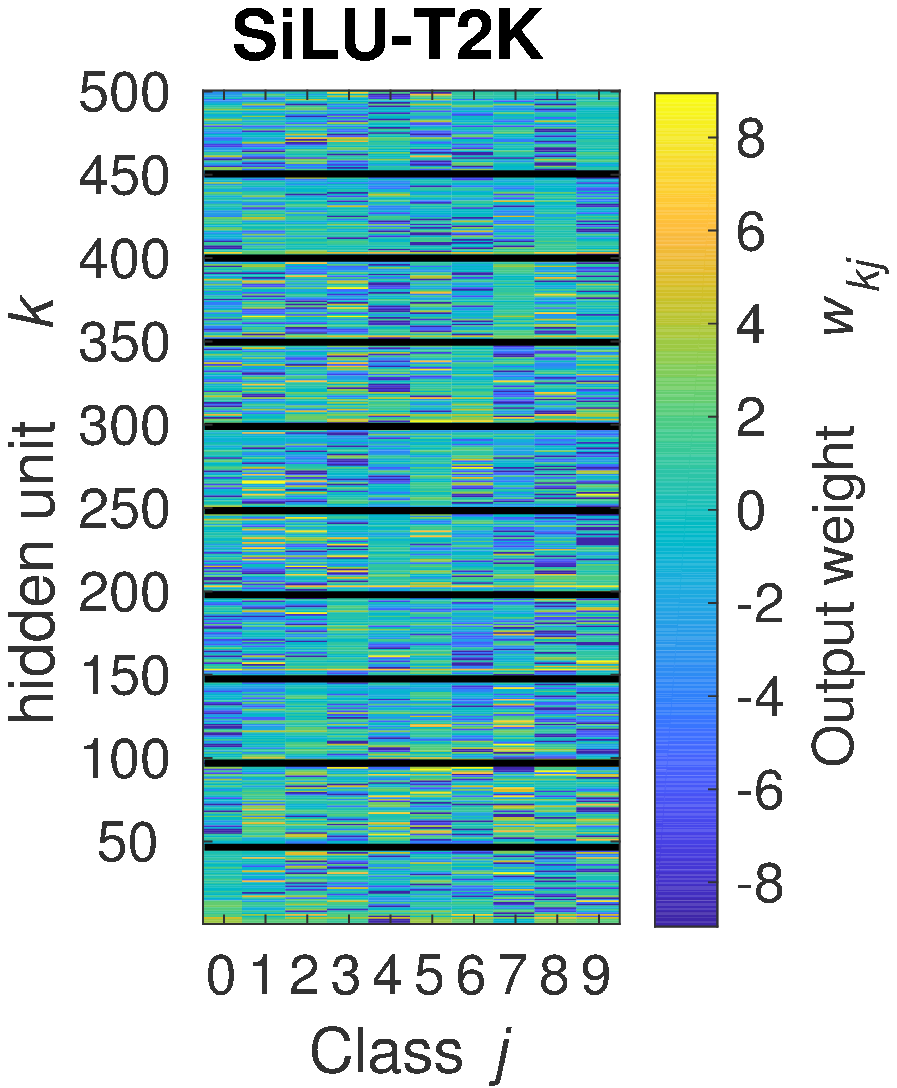}
}
\subfigure {
  \includegraphics[width=1.0\linewidth]{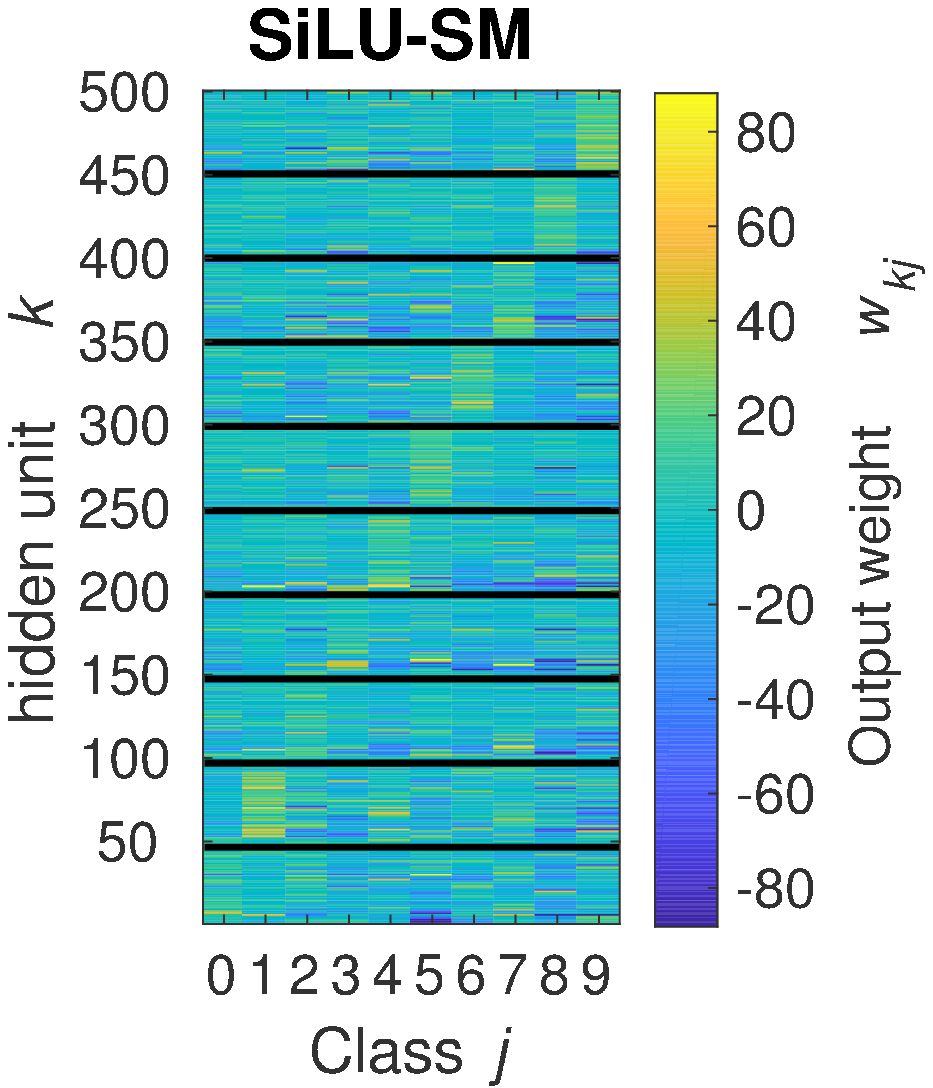}
} 
\caption{The learned output weights for the SiLU-T2K (top panel) and the
  SiLU-SM (bottom panel) networks. For each network, the hidden
  units were sorted by class according to the maximum value of the
  median activation and then grouped by class. The figure shows the
  output weights connected to the 50 hidden units with the highest
  median activation in each group}
\label{fig:EE_SM_w_out}
\end{center}
\end{figure} 
To further investigate the difference between the two methods, we
looked at the weights in the output
layers. Fig.~\ref{fig:EE_SM_w_out} shows the values of the trained
weights in the output layer for the SiLU-T2K network and the SiLU-SM
network. The hidden units were sorted by class according to the
maximum value of the median activation and then grouped by class. The
figure shows the output weights connected to the 50 hidden units with
the highest median activation in each group. The visualized data shows
two obvious differences between the two methods. First, the range of
the trained SiLU-SM weights were about a magnitude larger than the
range of the trained SiLU-T2K weights. Second, the trained SiLU-SM
network had a less shared (or less global) output weight structure, as
shown by higher positive values (yellow colors) for the
rectangles along the diagonal and mostly values with smaller
magnitudes (greenish color) outside the diagonal. To a large degree,
the SiLU-SM network learned separate classifiers, using non-overlapping
subsets of the hidden units, for each class.

\section{Conclusion}
In this study, inspired by the EE-RBM, we proposed unbounded output networks, UBnets, with unbounded units in the output layer and where the target value for correct classification $T \gg 1$. The UBnets are trained by a modified mean-squared error training objective, which weighs errors corresponding to correct (incorrect) classification more (less). 

We demonstrated, using shallow UBnets on MNIST, that a setting of $T$ equal to the number of hidden units significantly outperformed a setting of $T = 1$ and it also outperformed the reported results of standard neural networks by about 25\%. Using unbounded output residual networks, UBRnets, we validated our approach by achieving high-level classification performance on the MNIST, CIFAR-10, and CIFAR-100 datasets. 

Finally, we used MNIST to demonstrate that UBnets are much more resilient against adversarial examples than the standard approach of using a softmax output layer and training the networks by a cross-entropy objective.

\ifCLASSOPTIONcaptionsoff
  \newpage
\fi

\bibliographystyle{IEEEtran}
\bibliography{IEEEabrv,classification}

\begin{thebibliography}{10}
\providecommand{\url}[1]{#1}
\csname url@samestyle\endcsname
\providecommand{\newblock}{\relax}
\providecommand{\bibinfo}[2]{#2}
\providecommand{\BIBentrySTDinterwordspacing}{\spaceskip=0pt\relax}
\providecommand{\BIBentryALTinterwordstretchfactor}{4}
\providecommand{\BIBentryALTinterwordspacing}{\spaceskip=\fontdimen2\font plus
\BIBentryALTinterwordstretchfactor\fontdimen3\font minus
  \fontdimen4\font\relax}
\providecommand{\BIBforeignlanguage}[2]{{%
\expandafter\ifx\csname l@#1\endcsname\relax
\typeout{** WARNING: IEEEtran.bst: No hyphenation pattern has been}%
\typeout{** loaded for the language `#1'. Using the pattern for}%
\typeout{** the default language instead.}%
\else
\language=\csname l@#1\endcsname
\fi
#2}}
\providecommand{\BIBdecl}{\relax}
\BIBdecl

\bibitem{Elfwing15}
S.~Elfwing, E.~Uchibe, and K.~Doya, ``Expected energy-based restricted
  boltzmann machine for classification,'' \emph{Neural Networks}, vol.~64,
  no.~3, pp. 29--38, 2015.

\bibitem{Elfwing16}
------, ``From free energy to expected energy: Improving energy-based value
  function approximation in reinforcement learning,'' \emph{Neural Networks},
  vol.~84, pp. 17--27, 2016.

\bibitem{Elfwing18}
\BIBentryALTinterwordspacing
------, ``Sigmoid-weighted linear units for neural network function
  approximation in reinforcement learning,'' \emph{Neural Networks}, 2018.
  [Online]. Available: \url{https://doi.org/10.1016/j.neunet.2017.12.012}
\BIBentrySTDinterwordspacing

\bibitem{Elfwing18b}
------, ``Online meta-learning by parallel algorithm competition,'' in
  \emph{Proceedings of the Genetic and Evolutionary Computation Conference
  ({GECCO})}, 2018, pp. 426--433.

\bibitem{Mnih15}
V.~Mnih, K.~Kavukcuoglu, D.~Silver, A.~A. Rusu, J.~Veness, M.~G. Bellemare,
  A.~Graves, M.~Riedmiller, A.~K. Fidjeland, G.~Ostrovski, S.~Petersen,
  C.~Beattie, A.~Sadik, I.~Antonoglou, H.~King, D.~Kumaran, D.~Wierstra,
  S.~Legg, and D.~Hassabis, ``Human-level control through deep reinforcement
  learning,'' \emph{Nature}, vol. 518, no. 7540, pp. 529--533, 2015.

\bibitem{Elfwing17a}
S.~Elfwing, E.~Uchibe, and K.~Doya, ``Sigmoid-weighted linear units for neural
  network function approximation in reinforcement learning,''
  \emph{arXiv:1702.03118 [cs.LG]}, 2017.

\bibitem{Ramachandran17}
P.~Ramachandran, B.~Zoph, and Q.~V. Le, ``Searching for activation functions,''
  \emph{arXiv:1710.05941 [cs.NE]}, 2017.

\bibitem{Nair10}
V.~Nair and G.~Hinton, ``Rectified linear units improve restricted boltzmann
  machines,'' in \emph{Proceedings of the International Conference on Machine
  Learning ({ICML})}, 2010, pp. 807--814.

\bibitem{Krizhevsky09}
A.~Krizhevsky, ``Learning multiple layers of features from tiny images,''
  University of Toronto, Tech. Rep., 2009.

\bibitem{Deng09}
J.~Deng, W.~Dong, R.~Socher, L.-J. Li, K.~Li, and L.~Fei-Fei, ``{ImageNet: A
  Large-Scale Hierarchical Image Database},'' in \emph{CVPR09}, 2009.

\bibitem{Szegedy13}
C.~Szegedy, W.~Zaremba, I.~Sutskever, J.~Bruna, D.~Erhan, I.~J. Goodfellow, and
  R.~Fergus, ``Intriguing properties of neural networks.''
  \emph{arXiv:1312.6199 [cs.CV]}, 2013.

\bibitem{Smolensky86}
P.~Smolensky, ``Information processing in dynamical systems: Foundations of
  harmony theory,'' in \emph{Parallel Distributed Processing: Explorations in
  the Microstructure of Cognition. Volume 1: Foundations}, D.~E. Rumelhart and
  J.~L. McClelland, Eds.\hskip 1em plus 0.5em minus 0.4em\relax MIT Press,
  1986.

\bibitem{Freund92}
Y.~Freund and D.~Haussler, ``Unsupervised learning of distributions on binary
  vectors using two layer networks,'' in \emph{Proceedings Advances in Neural
  Information Processing Systems ({NIPS})}, 1992.

\bibitem{Hinton02}
G.~E. Hinton, ``Training products of experts by minimizing contrastive
  divergence,'' \emph{Neural Computation}, vol.~12, no.~8, pp. 1771--1800,
  2002.

\bibitem{LeCun98}
Y.~LeCun, L.~Bottou, Y.~Bengio, and P.~Haffner, ``Gradient-based learning
  applied to document recognition,'' in \emph{Proceedings of the IEEE}, vol.
  86(11), 1998, pp. 2278--2324.

\bibitem{He16}
K.~He, X.~Zhang, S.~Ren, and J.~Sun, ``Deep residual learning for image
  recognition,'' in \emph{Proceedings of the Conference on Computer Vision and
  Pattern Recognition ({CVPR2016})}, 2016, pp. 770--778.

\bibitem{Ioffe15}
S.~Ioffe and C.~Szegedy, ``Batch normalization: Accelerating deep network
  training by reducing internal covariate shift,'' in \emph{Proceedings of the
  International Conference on Machine Learning ({ICML})}, 2015, pp. 448--456.

\bibitem{He16b}
K.~He, X.~Zhang, S.~Ren, and J.~Sun, ``Identity mappings in deep residual
  networks,'' in \emph{Proceeding of European Conference on Computer Vision
  ({ECCV})}, 2016, pp. 630--645.

\bibitem{Mishkin16}
D.~Mishkin and J.~Matas, ``All you need is a good init,'' in \emph{Proceedings
  of the International Conference on Learning Representations ({ICLR})}, 2016.

\bibitem{He15}
K.~He, X.~Zhang, S.~Ren, and J.~Sun, ``Delving deep into rectifiers: Surpassing
  human-level performance on imagenet classification,'' in \emph{Proceedings of
  the International Conference on Computer Vision ({ICCV})}, 2015.

\bibitem{Srivastava15}
R.~K. Srivastava, K.~Greff, and J.~Schmidhuber, ``Training very deep
  networks,'' in \emph{Proceedings of Advances in Neural Information Processing
  Systems ({NIPS})}, 2015, pp. 2377--2385.

\bibitem{Vedaldi15}
A.~Vedaldi and K.~Lenc, ``Matconvnet: Convolutional neural networks for
  matlab,'' in \emph{Proceedings of the ACM International Conference on
  Multimedia ({ACMMM})}, 2015.

\bibitem{Goodfellow13}
I.~J. Goodfellow, D.~Warde-Farley, M.~Mirza, A.~Courville, and Y.~Bengio,
  ``Maxout networks,'' in \emph{Proceedings of the international conference on
  machine learning ({ICML})}, 2013.

\bibitem{Mairal14}
J.~Mairal, P.~Koniusz, Z.~Harchaoui, and C.~Schmid, ``Convolutional kernel
  networks,'' in \emph{Proceedings of Advances in Neural Information Processing
  Systems ({NIPS})}, 2014, pp. 2627--2635.

\bibitem{Lee15}
C.~Lee, S.~Xie, P.~Gallagher, Z.~Zhang, and Z.~Tu, ``Deeply-supervised nets,''
  in \emph{Proceedings of International Conference on Artificial Intelligence
  and Statistics ({AISTATS})}, 2015.

\bibitem{Liang15}
M.~Liang and X.~Hu, ``Recurrent convolutional neural network for object
  recognition,'' in \emph{Proceedings of the Conference on Computer Vision and
  Pattern Recognition ({CVPR})}, 2015, pp. 3367--3375.

\bibitem{Lee16}
C.~Lee, P.~W. Gallagher, and Z.~Tu, ``Generalizing pooling functions in
  convolutional neural networks: Mixed, gated, and tree,'' in \emph{Proceedings
  of the International Conference on Artificial Intelligence and Statistics
  ({AISTATS})}, 2016, pp. 464--472.

\bibitem{Chang15}
J.~Chang and Y.~Chen, ``Batch-normalized maxout network in network,''
  \emph{arXiv:1511.02583 [cs.CV]}, 2015.

\bibitem{Springenberg14}
J.~T. Springenberg, A.~Dosovitskiy, T.~Brox, and M.~A. Riedmiller, ``Striving
  for simplicity: The all convolutional net,'' \emph{arXiv:1412.6806 [cs.LG]},
  2014.

\bibitem{Clevert16}
D.~A. Clevert, T.~Unterthiner, and S.~Hochreiter, ``Fast and accurate deep
  network learning by exponential linear units ({ELU}s),'' in \emph{Proceedings
  of the International Conference on Learning Representations ({ICLR})}, 2016.

\bibitem{Larsson17}
G.~Larsson, M.~Maire, and G.~Shakhnarovich, ``Fractalnet: Ultra-deep neural
  networks without residuals,'' in \emph{Proceedings of the International
  Conference on Learning Representations ({ICLR})}, 2017.

\bibitem{Huang16}
G.~Huang, Y.~Sun, Z.~Liu, D.~Sedra, and K.~Q. Weinberger, ``Deep networks with
  stochastic depth,'' in \emph{Proceeding of European Conference on Computer
  Vision ({ECCV})}, 2016, pp. 630--645.

\bibitem{Zagoruyko16}
S.~Zagoruyko and N.~Komodakis, ``Wide residual networks,'' in \emph{Proceedings
  of the British Machine Vision Conference ({BMVC})}, 2016.

\bibitem{Huang17}
G.~Huang, Z.~Liu, and K.~Q. Weinberger, ``Densely connected convolutional
  networks,'' \emph{arXiv:1608.06993 [cs.CV]}, 2017.

\bibitem{Simard03}
P.~Y. Simard, D.~Steinkraus, and J.~C. Platt, ``Best practices for
  convolutional neural networks applied to visual document analysis,'' in
  \emph{Proceedings of the International Conference on Document Analysis and
  Recognition ({ICDAR})}, 2003, pp. 958--963.

\bibitem{Srivastava14}
N.~Srivastava, G.~Hinton, A.~Krizhevsky, I.~Sutskever, and R.~Salakhutdinov,
  ``Dropout: A simple way to prevent neural networks from overfitting,''
  \emph{Journal of Machine Learning Research}, vol.~15, pp. 1929--1958, 2014.

\bibitem{Biggio13}
B.~Biggio, I.~Corona, D.~Maiorca, B.~Nelson, N.~{\v{S}}rndi{\'{c}}, P.~Laskov,
  G.~Giacinto, and F.~Roli, ``Evasion attacks against machine learning at test
  time,'' in \emph{Proceedings of the European Conference on Machine Learning
  and Knowledge Discovery in Databases ({ECML-KDD})}, 2013, pp. 387--402.

\bibitem{Goodfellow15}
I.~J. Goodfellow, J.~Shlens, and C.~Szegedy, ``Explaining and harnessing
  adversarial examples,'' in \emph{Proceedings of the International Conference
  on Learning Representations ({ICLR})}, 2015.

\end{thebibliography}

\end{document}